\documentclass[letterpaper, 10 pt, conference]{ieeeconf}  

\IEEEoverridecommandlockouts                              

\usepackage[OT1]{fontenc} 
\usepackage{graphicx} 
\usepackage{multirow}
\usepackage{colortbl}
\usepackage{algorithm}
\usepackage{algpseudocode}
\usepackage{amsmath}
\usepackage{amsfonts}
\usepackage{stfloats}
\usepackage{subfigure}
\usepackage{makecell}
\usepackage{float}
\usepackage{caption}
\usepackage{booktabs,makecell, multirow, tabularx}
\usepackage{threeparttable}

\makeatletter
\let\NAT@parse\undefined
\makeatother
\usepackage[colorlinks,linkcolor=blue,anchorcolor=blue]{hyperref}
\title{\LARGE \bf
	Automatic MILP Model Construction for Multi-Robot Task Allocation and Scheduling Based on Large Language Models
}

\author{Mingming Peng$^{1,\dagger}$, Zhendong Chen$^{1,\dagger}$, Jie Yang$^{1}$, Jin Huang$^{1}$, Zhengqi Shi$^{1}$, Qihao Liu$^{1}$, Xinyu Li$^{1}$, Liang Gao$^{1}$
	\thanks{*This work is supported in part by the National Key Research and Development Program of China under Grant No. 2023YFB4705000 and National Natural Science Foundation of China under Grant 52188102.}
	\thanks{$^{1}$M. Peng, Z. Chen, J. Yang, J. Huang, Z. Shi, Q. Liu, X. Li, L. Gao are with the State Key Laboratory of Intelligent Manufacturing Equipment and Technology,  Huazhong University of Science and Technology, China.}%
        \thanks{$\dagger$ Equally contributed to this work.}
	\thanks{Correspondence: Xinyu Li, {\tt\small lixinyu@hust.edu.cn}.}%
}

\begin{document}
\maketitle	
\begin{abstract}
With the accelerated development of Industry 4.0, intelligent manufacturing systems increasingly require efficient task allocation and scheduling in multi-robot systems. However, existing methods rely on domain expertise and face challenges in adapting to dynamic production constraints. Additionally, enterprises have high privacy requirements for production scheduling data, which prevents the use of cloud-based large language models (LLMs) for solution development. To address these challenges,  there is an urgent need for an automated modeling solution that meets data privacy requirements.
 This study proposes a knowledge-augmented mixed integer linear programming (MILP) automated formulation framework, integrating local LLMs with domain-specific knowledge bases to generate executable code from natural language descriptions automatically. The framework employs a knowledge-guided DeepSeek-R1-Distill-Qwen-32B model to extract complex spatiotemporal constraints (82\% average accuracy) and leverages a supervised fine-tuned Qwen2.5-Coder-7B-Instruct model for efficient MILP code generation (90\% average accuracy). Experimental results demonstrate that the framework successfully achieves automatic modeling in the aircraft skin manufacturing case while ensuring data privacy and computational efficiency. This research provides a low-barrier and highly reliable technical path for modeling in complex industrial scenarios.
	
\end{abstract}
	
	\section{Introduction}

	With the advancement of Industry 4.0, intelligent manufacturing has emerged as a cornerstone of global manufacturing evolution. Multi-robot systems are extensively deployed in flexible production\cite{c1}\cite{c2}\cite{c3}, intelligent warehousing\cite{c4}, and logistics\cite{c5}. For instance, aircraft skin processing involves sequential processes such as material handling, bending, riveting, and adhesive application, each requiring coordination among heterogeneous robots\cite{c6}. Critical temporal constraints exist between adhesive application and riveting: the adhesive’s chemical properties degrade over time, necessitating riveting within its optimal activity window to ensure structural integrity. Similarly, welding requires post-process cooling before grinding. Dynamic production changes, such as prioritized skin processing for urgent deliveries, further complicate scheduling under strict time windows.
    
	The study of multi-robot task allocation and scheduling falls within the research domain of multi-robot task allocation (MRTA). According to the classification framework established by Gerkey et al.~\cite{c7}~\cite{c8}, the research subject of this paper can be precisely categorized under the single-task robot, single-robot task, time-extended assignment (ST-SR-TA) problem type. Existing research includes the Tercio algorithm by Gombolay et al.~\cite{c3}, which treats humans and robots as collaborative agents for task allocation and scheduling in manual operations. However, Ham et al.~\cite{c9} found that the Tercio algorithm outperforms MILP in terms of time, but its solution is more than 10\% away from the optimal in large-scale instances. They proposed a CP model for Boeing 777 manufacturing tasks. Guo ~\cite{c1} further incorporated logistics considerations into MILP and CP models for Boeing and Airbus manufacturing. Raatz et al.~\cite{c10} modeled humans and robots as unique resources and defined the problem as a flexible job shop scheduling problem (FJSP), solving it with a genetic algorithm. However, their approach lacks neighborhood optimization for critical paths, making it challenging to achieve optimality in large-scale instances\cite{c11}. In addition, these methods require domain-specific knowledge and frequent model adjustments for dynamic production changes. 
	
	In recent years, the emergence of large language models (LLMs) have opened new avenues for addressing complex combinatorial optimization problem~\cite{c12}. Current LLM applications in this domain primarily focus on automated model building and rule evolution. For instance, LLMs' generative capabilities enable the automatic building of mathematical models, effectively reducing manual modeling efforts and associated errors. Notably, successful integrations of LLMs with evolutionary algorithms have been demonstrated in solving traveling salesman problems (TSP)~\cite{c13}\cite{c14} and online bin packing problems (BPP)~\cite{c12}. However, when applied to workshop scheduling scenarios, existing rule evolution methods still exhibit substantial deviations from optimal solutions~\cite{c15}. Therefore, for aircraft skin processing workshops with a limited number of tasks, how to use large models to automatically construct MILP models becomes the focus of this study.

        \begin{figure}[!ht]
	\centering
	\includegraphics[width=8cm]{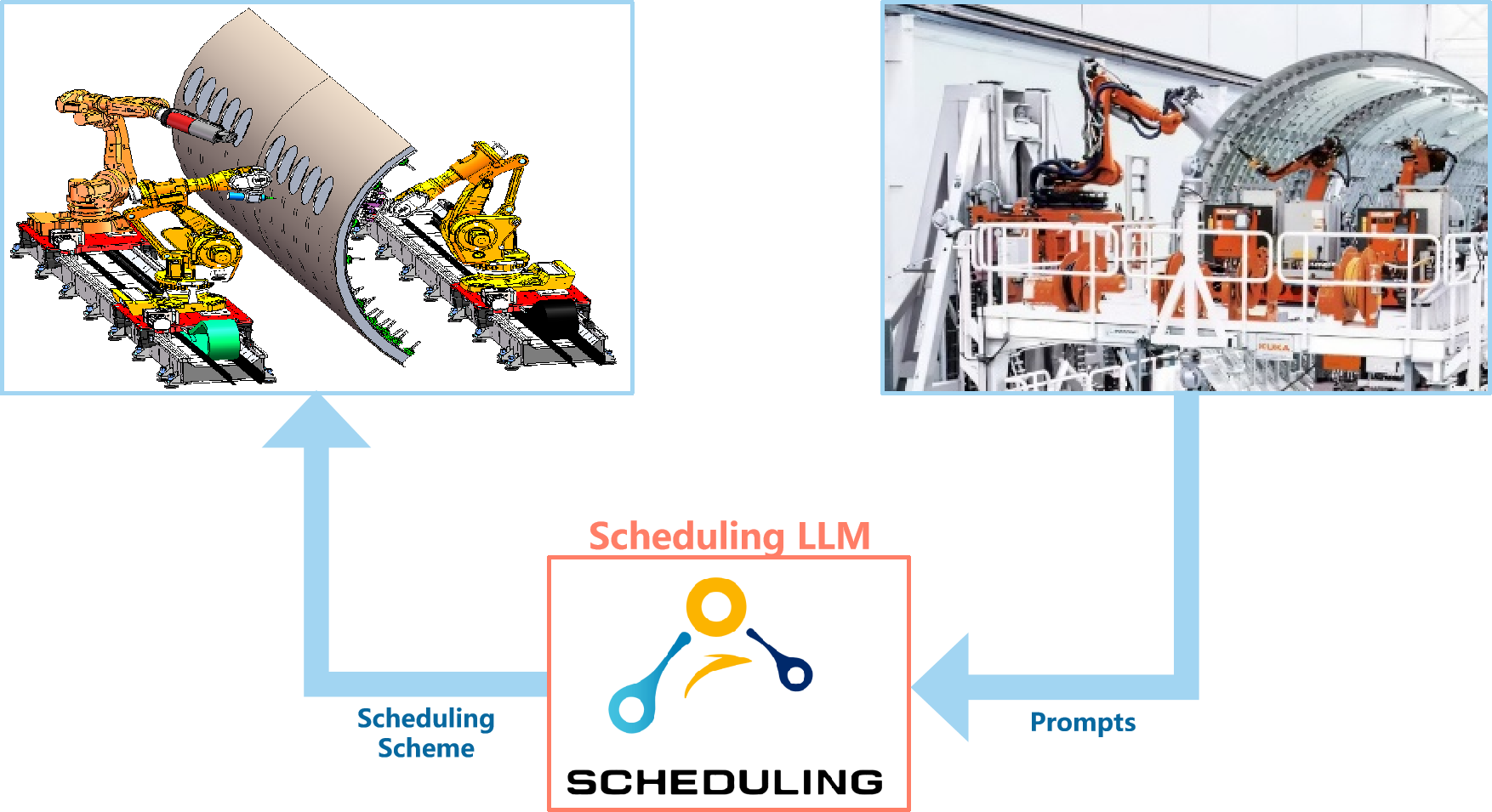}			
	\caption{LLM-Augmented Framework for Multi-Robot Task Allocation and Scheduling.}
	\label{fig1}
        \end{figure}

	
	For LLM-automated MILP building, Xiao et al.~\cite{c16} enhanced Operations Research (OR) problem reasoning through Chain-of-Experts, achieving validation on linear programming word problems (LPWP) and ComplexOR datasets albeit with limited formulation accuracy and reliance on cloud-based LLM APIs. AhmadiTeshnizi et al.~\cite{c17} employed modular architectures with connectivity graphs to process lengthy problem descriptions. Ye et al.~\cite{c18} integrated LLMs with large neighborhood search algorithms for solving MILP problems, employing a dual-layer self-evolutionary LLM agent to automate neighborhood strategy selection. While reducing expert knowledge dependency, this method exhibits two limitations: (1) interpretability deficits in generated strategies and (2) constrained generalization in robotic smart manufacturing scheduling due to intelligent training datasets. Li et al. ~\cite{c19} generated thousand-level problem classes through LLM evolution to establish the first MILP foundation model. However,  practical applications revealed defects, including problem class evolution path dependency on seed data and limited complex constraint generation capabilities.\par
    
    Current mainstream solutions rely on cloud-based LLM API calls ~\cite{c16}~\cite{c17}. However, cloud-based LLMs suffer from the disadvantages of insufficient domain expertise and high inference costs. Tests have shown that when cloud-based LLMs are not equipped with a workshop scheduling knowledge base, their success rate in formulating and solving the multi-robot task allocation and scheduling problem is relatively low. At the same time, enterprises have extremely high privacy requirements regarding production scene data and scheduling data. Uploading such data to cloud models poses a significant risk of privacy leakage. Therefore, enterprises require local LLMs capable of autonomously formulating MILP models and supporting local deployment. However, existing local LLMs exhibit poor analysis and reasoning capabilities, lack domain-specific knowledge, and fail to meet enterprise requirements regarding modeling and solution accuracy. AhmadiTeshnizi et al. ~\cite{c17} tested the automatic design of MILP mathematical models using an 8B LLM, with the test results showing an accuracy rate of less than 8\%. Additionally, the mathematical model used was not even the more complex workshop scheduling model. However, once the production workshop of a company is determined, its underlying mathematical model remains largely unchanged, with variations mainly occurring in factors such as the number of robots and processing constraints. This opens up possibilities for automatically formulating and solving MILP mathematical models based on LLMs.

    To overcome these challenges, we propose a knowledge-augmented automated MILP formulation framework, which systematically divides the construction process into two phases. First, we establish an FJSP model vector dataset and formulate flexible intelligent scheduling models for multi-robot systems using our local DeepSeek-R1-Distill-Qwen-32B model guided by domain knowledge. Subsequently, we fine-tune the locally deployed Qwen2.5-Coder-7B-Instruct model to enable automatic code generation based on mathematical formulations. The principal contributions of this work include:  
    \begin{itemize}
    \item[1)] A knowledge-augmented framework that automates the construction of six critical aircraft skin workshop constraints through integration with our local DeepSeek-R1-Distill-Qwen-32B model, demonstrating 82\% modeling accuracy. \par
    \item[2)] Development of a prompt-engineered formula-to-code translation dataset that enhances the Qwen2.5-Coder-7B-Instruct model's code generation accuracy from MILP formulations from about 5\% to 90\%.\par
    \item[3)] The proposed framework achieves cost-effective and lightweight automated formulation and solution of the multi-robot task allocation and scheduling problem.
    \end{itemize}\par

	\section{PROBLEM STATEMENT} \label{PROBLEM STATEMENT}

	The multi-robot task allocation and scheduling problem involves assigning multiple tasks to multiple robots, where each task comprises subtasks with multiple robot options. The objective is to assign optimal robots to subtasks and determine their processing sequences to minimize the makespan (maximum completion time). This problem extends the FJSP and inherits its constraints:
    
    \textbf{Initial state:} All robots are available at time 0 and all jobs are processable at time 0.
    
    \textbf{Robot occupancy:} A robot can process at most one subtask at any time.
    
    \textbf{Subtask processing:} A subtask can only be processed by one robot at a time, and subtasks within the same task follow predefined sequences, whereas subtasks across tasks are independent.
    
    \textbf{Non-preemption:} Once started, a subtask cannot be interrupted.
    
    \textbf{Other constraints:} Transition times between tasks on the same robot and transportation times between robots are negligible. The processing times are predefined.

    Additional scenario-specific constraints include:

    \textbf{Release time:} Task \(i\) can start no earlier than time \(t\). 
    
    \textbf{Deadline:} Task \(i\) must finish by time \(t\).
    
    \textbf{Time window:} Task \(i\) must start between \(t_1\) and \(t_2\).
    
    \textbf{Minimum interval:} After subtask \(j\) of task \(i\), a \(t\)-unit inspection delay is required before the next subtask.
    
    \textbf{Maximum interval:} The next subtask after subtask \(j\) of task \(i\) must start within \(t\) units.

    \textbf{Synchronization:} Tasks \(i_1\) and  \(i_2\) must complete simultaneously.

	\section{METHODOLOGY} \label{METHODOLOGY}

\begin{figure*}[!ht]
	\centering
	\includegraphics[width=13.5cm]{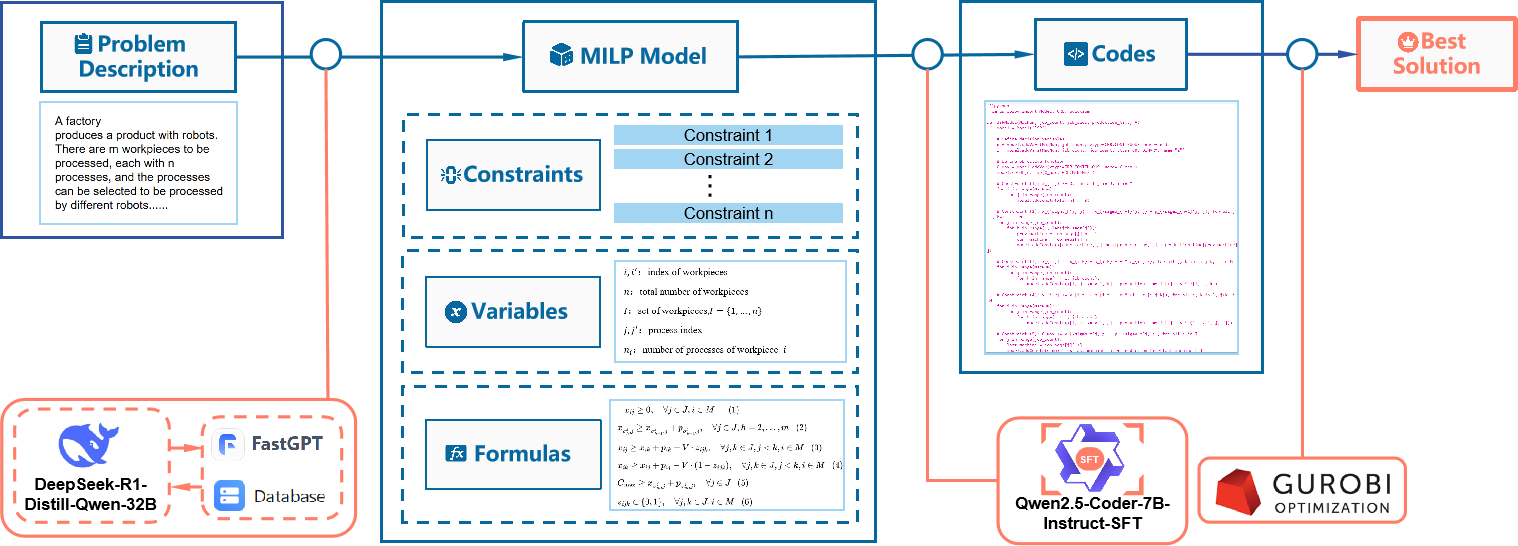}			
	\caption{The knowledge-augmented automated MILP formulation framework.}
	\label{fig2}
        \end{figure*}

	This study proposes a knowledge-augmented automated MILP formulation framework. The workflow comprises two phases: (1) translating fuzzy problem descriptions into precise MILP models and (2) converting MILP models into executable code. The framework is illustrated in Figure~\ref{fig2}.

    1) Fuzzy Description to MILP Model:
    
    Using a knowledge base, local LLMs translate natural language descriptions into standardized constraints.  Mathematical relationships are derived to generate LaTeX-formulated variables and equations. Semantic deviations are corrected via knowledge base validation, ensuring complete MILP model generation.
    
    2) MILP Model to Code:
    
    A template-based prompt framework fills MILP variables and equations into code templates. Then, a supervised fine-tuned (SFT)~\cite{c20} local LLM converts the MILP model into executable code.  Finally, the Gurobi solver is used to solve the code. With its efficient linear programming and integer programming algorithms, Gurobi quickly verifies the correctness and validity of the model.

	\subsection{Model Selection and Knowledge Base Construction}

    Local LLMs are compact open-source models constructed through LLMs quantization technology, including GLM4-9b-chat, Qwen2.5-Coder-7B-Instruct, Meta-Llama-3-8B-Instruct, and DeepSeek-R1-Distill-Qwen-32B. These models enable localized deployment, effectively controlling operational costs while ensuring data privacy protection. In practical applications, local LLMs leverage pre-constructed knowledge bases and employ natural language processing (NLP) techniques—particularly text classification and semantic understanding algorithms—to accurately identify the types of scheduling problems based on natural language inputs. This process facilitates building of model frameworks.
    
    The knowledge base adopts a hybrid architecture integrating LLMs with localized knowledge graphs. Advanced technologies such as vector indexing and knowledge graph embedding are implemented to enable efficient organization, storage, and retrieval of domain-specific knowledge. Configuration and management of the knowledge base are accomplished through the FastGPT platform, which enhances knowledge retrieval recall and accuracy via its state-of-the-art text generation and semantic understanding engines. After configuration, models are deployed and made accessible through remote API calls, ensuring system efficiency and scalable expansion.
    
    To equip LLMs with foundational knowledge for solving multi-robot task allocation and scheduling problems, we systematically organize basic MILP formulations of FJSP (excluding complex scenario constraints) and import them into the FastGPT platform for knowledge base construction. The formulation is presented as follows:
    
    1) Problem Description:
    
    Consider $n$ jobs to be processed on $m$ machines. Each job $i$ consists of $n_i$ operations, and operation $O_{i,j}$ can be selected to be processed on $m_{ij}$ machines. The scheduling objective involves assigning each operation to its most suitable machine and determining the optimal processing sequence for all operations on each machine to minimize the makespan.
    
    2) Variable definition:

\begin{itemize}
    \item $i, i'$: Job index
    \item $n$: Total number of jobs
    \item $I$: The set of jobs, $I = \{1, \ldots, n\}$
    \item $j, j'$: Operation index
    \item $n_i$: Number of operations for job $i$
    \item $J_i$: The set of operations for job $i$, $J_i = \{1, 2, \ldots, n_i\}$
    \item $O_{i,j}$: The $j$-th operation of job $i$
    \item $k$: Machine index
    \item $m$: Total number of machines
    \item $K$: Machine set, $K = \{1, 2, \ldots, m\}$
    \item $m_{ij}$: Available machine number for $O_{ij}$
    \item $K_{ij}$: Available machine set for $O_{ij}$
    \item $pt_{i,j,k}$: Processing time of $O_{ij}$ on machine $k$
    \item $M$: An extremely large positive number
    \item $X_{i,j,k}$: 0-1 decision variable, $X_{i,j,k} = 1$ when $O_{ij}$ is selected to be processed on machine $k$, otherwise $X_{i,j,k} = 0$
    \item $Y_{i,i,i',j'}$: $Y_{i,i,i',j'} = 1$ when $O_{ij}$ is processed before $O_{i',j'}$, otherwise $Y_{i,i,i',j'} = 0$
    \item $B_{i,j}$: Continuous decision variable indicating the start of processing time of $O_{ij}$
    \item $C_{\max}$: Makespan
\end{itemize}

    3) MILP model:
    
    The objective function is to minimize the makespan as shown in equation (1):
\begin{equation}
\min C_{\max} 
\end{equation}
    The MILP model needs to satisfy constraints (2)-(7).
\begin{equation}
\sum_{k \in K_{i, j}}  X_{i, j, k}=1, \forall i \in I, j \in J_i
\end{equation}
\begin{equation}
\begin{split}
B_{i, j}+ & \sum_{k \in K_{i, j}} X_{i, j, k} pt_{i, j, k} \\
&\leq B_{i, j+1}, \forall i \in I, j \in\left\{1, \ldots, n_i-1\right\}
\end{split}
\end{equation}
\begin{equation}
\begin{split}
B_{i, j}+ &  X_{i, j, k} p t_{i, j, k} \leq B_{i^{\prime}, j^{\prime}}+ \\
&M\left(3-Y_{i, j, i^{\prime}, j^{\prime}}- X_{i, j, k}- X_{i^{\prime} j^{\prime}, k}\right) \\
&\forall i \in I, i^{\prime} \in I, i<i^{\prime}, j \in J_i, j^{\prime} \in J_{i^{\prime},} k \in K_{i, j} \cap K_{i^{\prime}, j^{\prime}}
\end{split}
\end{equation}
\begin{equation}
\begin{split}
B_{i^{\prime}, j^{\prime}}+ &  X_{i^{\prime}, j^{\prime}, k} p t_{i^{\prime}, j^{\prime}, k} \leq B_{i, j}+ \\
&M\left(2+Y_{i, j, i^{\prime}, j^{\prime}}- X_{i, j, k}- X_{i^{\prime} j^{\prime}, k}\right) \\
&\forall i \in I, i^{\prime} \in I, i<i^{\prime}, j \in J_i, j^{\prime} \in J_{i^{\prime},} k \in K_{i, j} \cap K_{i^{\prime}, j^{\prime}}
\end{split}
\end{equation}
\begin{equation}
B_{i, n_i}+\sum_{k \in K_{i, n_i}} X_{i, n_i, k} p t_{i, n_i, k} \leq C_{\max }, \forall i \in I
\end{equation}
\begin{equation}
B_{i, j} \geq 0, \forall i \in I, j \in J_i
\end{equation}\par

 Constraint (2) indicates that any operation of any job can only select one machine from its available machine set. Constraint (3) specifies that different operations of the same job must satisfy the designated precedence processing order. Constraints (4) and (5) determine the sequential processing order between different operations scheduled on the same machine. Constraint (6) ensures that the makespan is no less than the completion time of any job. Constraint (7) enforces the non-negativity of operation start times.
    
\subsection{Data Generation}
\begin{figure}[!ht]
\centering
\includegraphics[width=8cm]{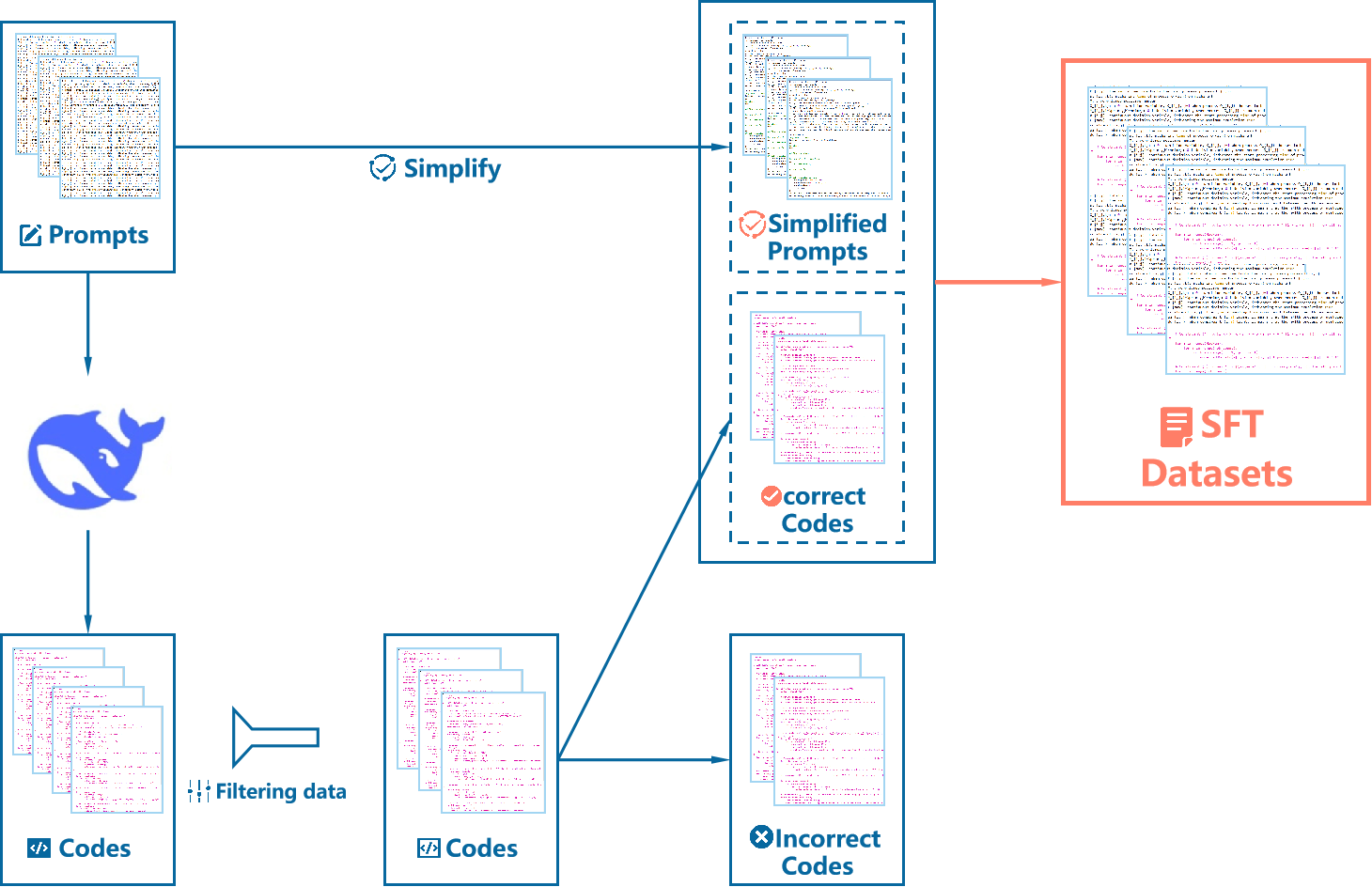}			
\caption{SFT dataset construction pipeline.}
\label{fig3}
\end{figure}

To enhance the local LLMs' capability in translating mathematical formulations into executable code, we perform SFT on them. The dataset production process is shown in Figure~\ref{fig3}. The workflow begins with organizing standard MILP formulations for combinatorial optimization problems and designing prompts to test the local LLMs. Finallly, the code generated by these models is submitted to the Gurobi solver for verification. Based on solver feedback (including solution results and error messages), we iteratively optimize the prompts until correct outputs are obtained. Notably, critical model information is intentionally included in prompts during this phase to improve accuracy but later removed during dataset construction to ensure generalization.

After acquiring multiple validated prompts, we conduct batch testing to generate extensive code data. This data underwent rigorous cleaning (including deduplication and runtime verification) to separate correct from erroneous code. Finally, we simplify complex prompts by removing verbose descriptions and key information. The simplified prompts and corresponding validated code are paired as instruction-output pairs to construct the SFT dataset. Through this methodology, over 10,000 data entries are collected, with 494 verified correct entries retained for dataset preparation after processing.
	
\subsection{Fine-Tuning Strategy}
We select LLaMA-Factory~\cite{c21}\cite{c22} as the fine-tuning platform for local LLMs due to its significant streamlining of the fine-tuning workflow. In this study, we implement low-rank adaptation (LoRA) for SFT of local LLMs. The LLaMA-Factory platform supports advanced training methodologies like LoRA, providing flexible fine-tuning strategies while maintaining computational efficiency. Crucially, the platform comprehensively logs critical training metrics and dynamically generates loss curves through real-time synchronization during training sessions. This capability enables precise monitoring and quantitative evaluation of model performance throughout optimization.

\section{EXPERIMENTAL RESULTS AND ANALYSIS} \label{EXPERIMENTAL RESULTS AND ANALYSIS}
This section evaluates the formulation capabilities and code generation performance of various models, selects two optimal LLMs, and validates the effectiveness of our proposed knowledge-augmented automated MILP formulation framework through an aircraft skin manufacturing case study.
    
\subsection{LLM Performance Benchmarking}
The local LLMs selected for the experiment are GLM4-9b-chat, Qwen2.5-Coder-7B-Instruct, Meta-Llama-3-8B-Instruct, and DeepSeek-R1-Distill-Qwen-32B, all of which are deployed on an in-house server.

\textbf{Implementation Details:} The knowledge base configuration platform utilized FastGPT v4.8.11, while the deployment and fine-tuning of local LLMs are supported by a hardware configuration of 4×NVIDIA GeForce RTX 4090D GPUs. During the formulation-to-code translation phase, the generated code is executed and validated in Visual Studio Code 1.97, with a runtime environment of Python 3.10.15 and the Gurobi 10.0.3 solver.
    
\subsubsection{Formulation Capability Analysis with Knowledge Base}
During the formulation phase, we evaluate the modeling capabilities of different local LLMs in handling various complex scenario constraints for the multi-robot task allocation and scheduling problem under knowledge base-supported conditions.

Analysis of experimental results (Table \ref{tab:addlabel5}) demonstrates significant disparities in constraint formulation capabilities among local LLMs under knowledge-augmented conditions. DeepSeek-R1-Distill-Qwen-32B, leveraging knowledge distillation techniques to inherit powerful constraint handling capabilities from cloud-based LLMs, achieves high formulation accuracy in most scenarios. Other local LLMs exhibit lower compliance rates in knowledge-guided constraint formulation, revealing inherent limitations in constraint formulation proficiency among local LLMs.



\begin{table}[htbp]
  \centering
  \caption{Formulation accuracy of different local LLMs with knowledge base}
  \label{tab:addlabel5}
  \begin{tabular}{m{5.5em}*{3}{m{5.5em}}}
    \toprule
    \cmidrule{1-4}
    & \textbf{Qwen2.5-7B-Instruct} 
      & \textbf{Meta-Llama-3-8B-Instruct} & \textbf{DeepSeek-R1-Distill-Qwen-32B} \\
    \midrule
    constraint1 & 50\%  & 80\%  & \textbf{100\% }\\
    constraint2 & 30\%  & 10\%  & \textbf{100\% }\\
    constraint3 & 90\%  & 70\%  & \textbf{100\%} \\
    constraint4 & 0\%   & 0\%   & \textbf{50\%} \\
    constraint5 & 0\%   & 0\%   & \textbf{60\%} \\
    constraint6 & 0\%   & 30\%  & \textbf{80\%} \\
    Average     & 28\%  & 32\%  & \textbf{82\%} \\
    \bottomrule
  \end{tabular}%
\end{table}

\begin{table}[htbp]
  \centering
  \caption{Code generation accuracy of different local LLMs after supervised fine-tuning}
  \label{tab:addlabel3}
  \begin{tabular}{m{5.5em}*{3}{m{5.5em}}}
    \toprule
    \cmidrule{1-4}
    & \textbf{Meta-Llama-3-8B-Instruct} 
      & \textbf{DeepSeek-R1-Distill-Qwen-7B} & \textbf{Qwen2.5-Coder-7B-Instruct} \\
    \midrule
    constraint1 & \textbf{100\%} & 90\%  & \textbf{100\%} \\
    constraint2 & 60\%  & \textbf{100\%} & 70\%  \\
    constraint3 & 10\%  & 60\%  & \textbf{100\%} \\
    constraint4 & 0\%   & 80\%  & \textbf{100\%} \\
    constraint5 & 20\%  & 0\%   & \textbf{70\%}  \\
    constraint6 & 0\%   & 0\%   & \textbf{100\%} \\
    Average     & 32\%  & 55\%  & \textbf{90\%}  \\
    \bottomrule
  \end{tabular}%
\end{table}

\subsubsection{Code Generation Accuracy under SFT}
During the MILP model-to-code phase, we compare the capability of different SFT local LLMs to convert validated MILP formulations into executable code. Utilizing the SFT dataset constructed for FJSP models through the methodology described in Section III,  SFT is applied to local LLMs. Through iterative adjustments of model parameters and training strategies, the local LLMs are optimized to better capture the structural patterns and operational logic inherent in FJSP formulations.

To validate code correctness, we design a benchmark case where the generated code is submitted to the Gurobi 10.0.3 solver for execution. 

Analysis of experimental results (Table \ref{tab:addlabel3}) demonstrates that local LLMs implementing SFT on fundamental FJSP datasets exhibit significant capability enhancement in MILP model-to-code conversion tasks. The SFT-enhanced Qwen2.5-Coder-7B-Instruct model achieves 100\% accuracy rates across 4 constraints (Constraints 1/3/4/6), while showing particularly notable performance in deadline constraints (Constraint 2) and maximum time-interval constraints (Constraint 5) with 70\% accuracy - significantly outperformed comparable models of similar scales (e.g., 10-20\% accuracy for Meta-Llama-3-8B-Instruct).

\subsection{Case Study: Aircraft Skin Manufacturing}
\begin{figure*}[!ht]
\centering
\includegraphics[width=13.5cm]{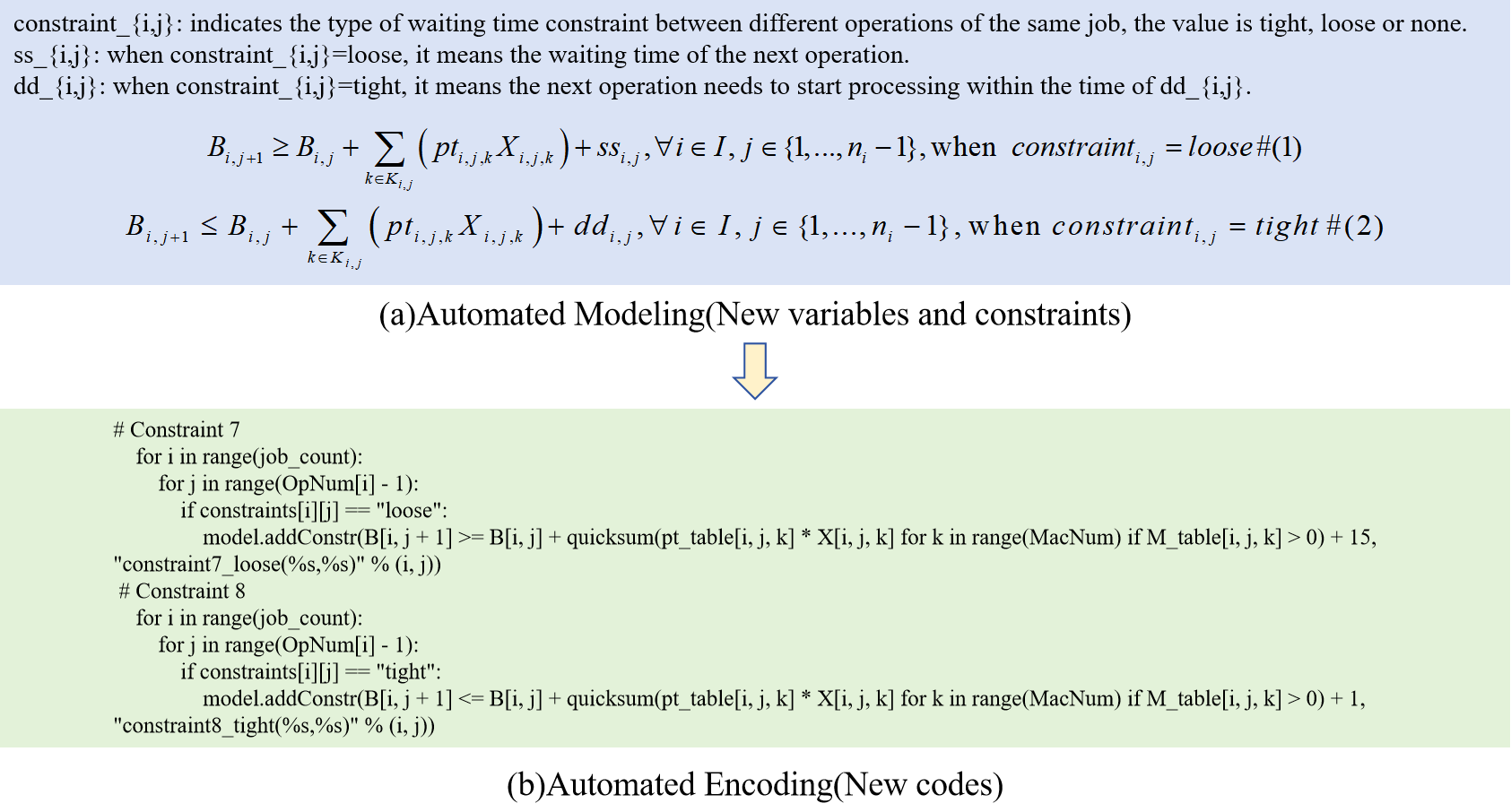}			
\caption{Result for cases automatically generated by knowledge-augmented automated MILP formulation framework.}
\label{fig4}
\end{figure*}

In aircraft skin manufacturing, complex operational constraints include rigorous temporal windows (e.g., riveting must occur within adhesive activation periods post-coating) and physical property requirements (mandatory cooling phases post-welding). We apply the knowledge-augmented automated MILP formulation framework to automatically formulate optimization models for this scenario. Based on the experimental evaluation of the previous, two local LLMs—DeepSeek-R1-Distill-Qwen-32B (with knowledge base integration) and SFT Qwen2.5-Coder-7B-Instruct are selected to autonomously construct MILP formulations for aircraft skin manufacturing. The generated code is then executed using the Gurobi solver to obtain production-optimized scheduling solutions.(Fig~\ref{fig4})

\begin{table*}[htbp]
  \centering
  \caption{Verification of automatically generated code with 10 different  examples}
  \begin{tabular}{cccccc|ccccc}
    \toprule
    Scales & \multicolumn{5}{c}{10 tasks and 5 robots}    & \multicolumn{5}{c}{10 tasks and 10 robots} \\
    \cmidrule(r){2-6} \cmidrule(l){7-11} 
    Cases & la01  & la02  & la03  & la04  & la05  & la06  & la07  & la08  & la09  & la10 \\
    makespan & 628   & 655   & 568   & 598   & 503   & 892   & 708   & 847   & 799   & 879 \\
    best solution & 628   & 655   & 568   & 598   & 503   & 892   & 708   & 847   & 799   & 879 \\
    \bottomrule
  \end{tabular}%
  \label{tab:addlabel4}%
\end{table*}%

Domain experts first validated the models, confirming the system's accurate identification of release time constraints and deadline constraints, along with the establishment of correct MILP formulations. Subsequently, 10 cases are designed to verify the automatically generated code, with a solver time limit set to 100 seconds. As shown in Table \ref{tab:addlabel4}, the obtained optimal solutions demonstrate successful code validation, thereby proving the operational effectiveness of the knowledge-augmented automated MILP formulation framework.
	
\section{CONCLUSIONS}	\label{conclusions}
This study proposes a knowledge-guided automated MILP modeling framework, which integrates NLP with domain knowledge base technology to achieve intelligent modeling of multi-robot task allocation and scheduling problems. The framework innovatively constructs a two-step transformation mechanism, which sequentially converts natural language into an MILP model and then translates the MILP model into executable code. First, DeepSeek-R1-Distill-Qwen-32B is used to parse vague problem descriptions and extract constraint semantics. Then, a knowledge-base-guided constraint validation mechanism ensures the integrity of complex relationships such as time-space and priority. Finally, the SFT Qwen2.5-Coder-7B-Instruct generates executable MILP model code. Experiments demonstrate that the proposed method performs well in modeling and solving multi-robot task allocation and scheduling problems. This result offers an automated modeling and solving solution for robotic intelligent manufacturing.

Current experiments have validated the feasibility of local LLMs in large-scale, complex constraint scenarios. The next phase will focus on enhancing computational efficiency and processing capability. Future work will focus on further improving the overall accuracy of automated mathematical model construction, while extending the problem to multi-robot task scheduling in a broader range of workshop scheduling scenarios.

\addtolength{\textheight}{-12cm}   





References are important to the reader; therefore, each citation must be complete and correct. If at all possible, references should be commonly available publications.


\begin{thebibliography}{99}

\bibitem{c1} D. Guo, "Fast scheduling of human-robot teams collaboration on synchronised production-logistics tasks in aircraft assembly," Robotics and Computer-Integrated Manufacturing, vol. 85, Elsevier, 2024, p. 102620.

\bibitem{c2} B. Altundas et al., "Learning coordination policies over heterogeneous graphs for human-robot teams via recurrent neural schedule propagation," in 2022 IEEE/RSJ International Conference on Intelligent Robots and Systems (IROS), 2022, pp. 11679--11686.


\bibitem{c3} M. C. Gombolay, R. J. Wilcox, and J. A. Shah, "Fast scheduling of robot teams performing tasks with temporospatial constraints," IEEE Transactions on Robotics, vol. 34, no. 1, IEEE, 2018, pp. 220--239.

\bibitem{c4} Z. Li et al., "A mechanism for scheduling multi robot intelligent warehouse system face with dynamic demand," Journal of Intelligent Manufacturing, vol. 31, Springer, 2020, pp. 469--480.

\bibitem{c5} A. Contini and A. Farinelli, "Coordination approaches for multi-item pickup and delivery in logistic scenarios," Robotics and Autonomous Systems, vol. 146, Elsevier, 2021, p. 103871.

\bibitem{c6} J. M. L. da Saúde and J. M. Silva, "Aircraft industrialization process: A systematic and holistic approach to ensuring integrated management of the engineering process," in Technology and Manufacturing Process Selection: The Product Life Cycle Perspective, Springer, 2013, pp. 81--103.

\bibitem{c7} B. P. Gerkey and M. J. Matarić, "A formal analysis and taxonomy of task allocation in multi-robot systems," The International journal of robotics research, vol. 23, no. 9, SAGE Publications, 2004, pp. 939--954.

\bibitem{c8} Q. Li et al., "Distributed near-optimal multi-robots coordination in heterogeneous task allocation," in 2020 IEEE/RSJ International Conference on Intelligent Robots and Systems (IROS), 2020, pp. 4309--4314.

\bibitem{c9} A. Ham and M.-J. Park, "Human--robot task allocation and scheduling: Boeing 777 case study," IEEE Robotics and Automation Letters, vol. 6, no. 2, IEEE, 2021, pp. 1256--1263.

\bibitem{c10} A. Raatz et al., "Task scheduling method for HRC workplaces based on capabilities and execution time assumptions for robots," CIRP Annals, vol. 69, no. 1, Elsevier, 2020, pp. 13--16.

\bibitem{c11} X. Li et al., "Improved gray wolf optimizer for distributed flexible job shop scheduling problem," Science China Technological Sciences, vol. 65, no. 9, Springer, 2022, pp. 2105--2115.

\bibitem{c12} B. Romera-Paredes et al., "Mathematical discoveries from program search with large language models," Nature, vol. 625, no. 7995, Nature Publishing Group UK London, 2024, pp. 468--475.

\bibitem{c13} F. Liu et al., "Algorithm evolution using large language model," arXiv preprint arXiv:2311.15249, 2023.


\bibitem{c14} F. Liu et al., "An example of evolutionary computation+ large language model beating human: Design of efficient guided local search," CoRR, 2024.


\bibitem{c15} J. Huang et al., "Automatic programming via large language models with population self-evolution for dynamic job shop scheduling problem," arXiv preprint arXiv:2410.22657, 2024.

\bibitem{c16} Z. Xiao et al., "Chain-of-experts: When llms meet complex operations research problems," in The twelfth international conference on learning representations, 2023.

\bibitem{c17} A. AhmadiTeshnizi, W. Gao, and M. Udell, "Optimus: Scalable optimization modeling with (mi) lp solvers and large language models," arXiv preprint arXiv:2402.10172, 2024.

\bibitem{c18} H. Ye et al., "Large Language Model-driven Large Neighborhood Search for Large-Scale MILP Problems," 2025.


\bibitem{c19} S. Li et al., "Towards foundation models for mixed integer linear programming," arXiv preprint arXiv:2410.08288, 2024.

\bibitem{c20} J. Li et al., "Getting more juice out of the sft data: Reward learning from human demonstration improves sft for llm alignment," Advances in Neural Information Processing Systems, vol. 37, 2025, pp. 124292--124318.

\bibitem{c21} Y. Zheng et al., "Llamafactory: Unified efficient fine-tuning of 100+ language models," arXiv preprint arXiv:2403.13372, 2024.

\bibitem{c22} Y. Kang and J. Kim, "ChatMOF: an artificial intelligence system for predicting and generating metal-organic frameworks using large language models," Nature communications, vol. 15, no. 1, Nature Publishing Group UK London, 2024, p. 4705.













\end{thebibliography}
\end{document}